\documentclass{article}

\usepackage{times}
\usepackage{graphicx} 
\usepackage{subfigure}
\usepackage{array}
\usepackage{float}
\usepackage{booktabs, caption}
\usepackage{natbib}
\usepackage{amsmath}
\usepackage{algorithm, algorithmicx, algpseudocode}

\usepackage[accepted]{icml2017}

\newtheorem{theorem}{Theorem}

\begin{document}

\icmltitlerunning{Convolution Aware Initialization}

\twocolumn[
\icmltitle{Convolution Aware Initialization}

\icmlsetsymbol{equal}{*}

\begin{icmlauthorlist}
	\icmlauthor{Armen Aghajanyan}{dim}
\end{icmlauthorlist}

\icmlaffiliation{dim}{DimensionalMechanics, Bellevue, Washington, United States}

\icmlkeywords{deep learning, initialization, convolution}

\vskip 0.3in
]

\begin{abstract}
	Initialization of parameters in deep neural networks has been shown to have a big impact on the performance of the networks \cite{DBLP:journals/corr/MishkinM15}. The initialization scheme devised by He et al, allowed convolution activations to carry a constrained mean which allowed deep networks to be trained effectively \cite{DBLP:journals/corr/HeZR015}. Orthogonal initializations and more generally orthogonal matrices in standard recurrent networks have been proved to eradicate the vanishing and exploding gradient problem \cite{DBLP:journals/corr/abs-1211-5063}. Majority of current initialization schemes do not take fully into account the intrinsic structure of the convolution operator. Using the duality of the Fourier transform and the convolution operator, Convolution Aware Initialization builds orthogonal filters in the Fourier space, and using the inverse Fourier transform represents them in the standard space. With Convolution Aware Initialization we noticed not only higher accuracy and lower loss, but faster convergence. We achieve new state of the art on the CIFAR10 dataset, and achieve close to state of the art on various other tasks.

\end{abstract}

\section{Introduction}
Deep neural networks have been extremely successful and have demonstrated impressive results in various structured data problems in fields such as computer vision and speech recognition \cite{Zagoruyko2016WRN, DBLP:journals/corr/SaonSRK16}.
One of the core building blocks of deep neural networks are layers that perform convolution in order to capture local structured information, whether it be two dimensional convolution for image classification or one dimensional convolutional for audio and NLP tasks \cite{DBLP:journals/corr/SpringenbergDBR14, DBLP:journals/corr/Kim14f}. A lot of work has been done on finding initialization schemas that allow for fast levels of convergence and good performance. The majority of initializations use the input and output dimensions of the convolutional layer to scale a specific distribution.

He et al describe a initialization schema where weights are sampled from a normal distribution with a scaled variance and zero mean in order to not have a exploding effect in activations of further layers \cite{DBLP:journals/corr/HeZR015}. Glorot initialization is a similar looking initialization schema defined by a different variance scaling factor \cite{Glorot10understandingthe}. Although these initialization schemes attempt to exploit some properties of convolution, the duality of the convolution operator and the Fourier transform have not been explored or exploited, to our knowledge.

The concept of orthogonality has been explored for initialization by various papers. Orthogonality has been shown to have numerous useful properties in both standard networks as well as recurrent networks \cite{DBLP:journals/corr/SaxeMG13, DBLP:journals/corr/MhammediHRB16}. Orthogonal initialization is beneficial due to several reasons. Orthogonal initialization produces stable matrices, in the sense that under repeated multiplication the matrix does not vanish or explode. This property has been exploited in recurrent networks to combat the vanishing and exploding gradient problem \cite{DBLP:journals/corr/SaxeMG13}. Orthogonality can also be reasoned to produce the most diverse set of features possible under a defined inner product space, one feature detector will capture information that would be completely missed by another feature detector.

Our algorithm uses the convolution theorem to represent the convolution operator in the product-sum space (frequency domain) where we then build our orthogonal representation. And utilizing the inverse Fourier transform we can represent that orthogonality in the standard space.

\section{Convolution Aware Initialization}
The main idea behind Convolution Aware Initialization (CAI) is in order to maximize the expressiveness power of convolutional layer, we form orthogonal filters not in the standard convolution space, but in the Fourier space. The reasoning is due to the convolutional theorem which states that convolution in the time domain is element wise multiplication in the frequency domain. The standard way to do orthogonal initialization in a convolution block is to flatten the 4 dimensional tensor into matrix, performing orthogonal decomposition and reshaping the respective matrix back into the correct sized tensor \cite{DBLP:journals/corr/SaxeMG13}. CAI defines the initialization scheme in a different manner. Below we define CAI for the 2-dimensional convolutional layer used commonly in image machine learning problems.

We can write a set of filters on an input space $x$ with convolutional operator $\otimes$, and filter $f_{i,j}=R^{k \times m}$ as:

\begin{equation}
	\{f_{0,j} \otimes x_0, f_{1,j} \otimes x_1, f_{2,j} \otimes x_2, ..., f_{n,j} \otimes x_n \}
\end{equation}
Across the stack of filters we reduce via a sum to achieve our singular output $j$.

\begin{equation}
	s_j = \sum_{i=0}^{n} f_{i,j} \otimes x_i
\end{equation}
We can apply the Fourier transform to exploit the convolution theorem.
\begin{equation}
	\mathcal{F}(s_j) = \sum_{i=0}^{n} \mathcal{F}(f_{i,j}) \odot \mathcal{F}(x_i)
\end{equation}
Since the filters and previous activation maps are in two dimensions the element wise multiplication can be thought of as Hadamard products. The previous activation maps recursively depend on previous filters and input, therefore manipulating the previous states into a expected state will allow us to rewrite as:

\begin{equation}
	\mathcal{F}(s_j) \oslash \mathrm{E}[\mathcal{F}(x)] = \sum_{i=0}^{n} c_i*\mathcal{F}(f_{i,j})
\end{equation}

We introduce a constant scaling $c_i$ which allows us to reason about the right side of the equation as a linear combination ($c_i$ can be thought of as a constant scaling factor that is pulled from $f_{i,j}$). Therefore the goal is to select the correct set of $\mathcal{F}(f_{i,j})$ to form a complete basis over the left side of the expression. Instead of forming an arbitrary basis we focus on forming an orthogonal basis using $\mathcal{F}(f_{i,j})$.

This can be done by building a matrix in $R^{\mathcal{F}_{km} \times n}$, diagonalizing, and taking the columns representing the eigenvectors and reshaping into $\mathcal{F}_{km}$, where $\mathcal{F}_{km}$ represents the size of the Fourier transformed matrix. The filters can then be transformed back from the standard domain using the inverse Fourier transform $\mathcal{F}^{-1}$.

For the sake of being detailed, the 2-dimensional Fourier transform used throughout the paper refers to a Fourier transform in the form of:

\begin{equation}
	A_{kl} = \sum_{m=0}^{M-1} \sum_{n=0}^{N-1} a_{mn} e^{-2\pi i \left[\frac{mk}{M} + \frac{nl}{N} \right] }
\end{equation}

With the inverse Fourier transform in the form of:
\begin{equation}
	a_{mn}  = \frac{1}{MN} \sum_{m=0}^{M-1} \sum_{n=0}^{N-1} A_{kl}  e^{2\pi i \left[\frac{mk}{M} + \frac{nl}{N} \right] }
\end{equation}

With indexes defined as:
\begin{align}
	k = \{0, ... M-1\} &   \\
	l = \{0, ... N-1\} &
\end{align}

\subsection{Properties of CAI}
\subsubsection{Bounds of CAI}

The magnitude of the post-decomposition matrix can be set using a scaling factor, but the question lies in what the magnitude of the filter weights will be after computing the inverse Fourier transform. If we define the inverse DFT, with inputs $(A_{kl})_{k=0, \ l=0}^{M-1, \ N-1}$, using the triangle inequality we get:

\begin{align}
	|\frac{1}{MN} \sum_{m=0}^{M-1} \sum_{n=0}^{N-1} A_{kl}  e^{2\pi i \left[\frac{mk}{M} + \frac{nl}{N} \right]} |       \\
	\leq \frac{1}{MN} \sum_{m=0}^{M-1} \sum_{n=0}^{N-1} | A_{kl}  e^{2\pi i \left[\frac{mk}{M} + \frac{nl}{N} \right]} | \\
	\leq \frac{1}{MN} \sum_{m=0}^{M-1} \sum_{n=0}^{N-1} | A_{kl} |
\end{align}

Therefore we can say that the range of the inverse Fourier transform is bounded by the average of the inputs magnitudes. This can be used in the future to scale the decomposition accordingly. We do not explicitly talk about the distribution sampled for the matrix that the decomposition is computed on. In reality it can be any distribution, but throughout this paper we sampled a normal distribution with a zero mean and one variance to construct a positive definite symmetric matrix.

\begin{theorem}
	Any square real  positive definite symmetric matrix $\mathbf{A}$ with a unique eigen-decomposition in the form of $\mathbf{Q}\mathbf{\Lambda}\mathbf{Q}^{-1}$ has an upper-bound of $1$ on $|\mathbf{Q}|$
\end{theorem}

Given vectors $x$ and $y$ and their corresponding eigenvalues $\lambda_1$ and $\lambda_2$ we can state $\langle \mathbf{A}x,y \rangle = \langle \mathbf{A}x,y \rangle$. It is trivial to show $\lambda_1 \langle x,y \rangle = \lambda_2 \langle x,y \rangle$ and therefore $(\lambda_1 -\lambda_2) \langle x,y \rangle$ proving that the eigenspaces of $\mathbf{Q}$ are orthogonal. Now we can find an orthonormal basis for each eigenspace and since the eigenspaces are mutually orthogonal, the vectors in the eigenspace form an orthonormal basis. We have shown that the vectors in $\mathbf{Q}$ form an orthonormal basis therefore the columns (or rows) must form a unit norm. If all vectors form a unit-norm individual entries in the matrix must have an upper-bound of 1 with respect to their magnitude.

Prior proof shows that the upper-bound for CAI prior to linear scaling will be 1.

\subsubsection{Expected Value of CAI}

He et al. derived the specific variance and mean needed in order to insure that the activations of the convolutional network will not explode. In our initialization scheme we correct the filters through linear scaling in a way to preserve the variance, while maintaining orthogonality of the filters in the Fourier space. The natural question is what the distribution CAI is, and more importantly where does the mean lie. If we assume that every element in the a matrix belongs to a single distribution per matrix, the expectation is the expectation of the distribution, and can be approximated by averaging all elements in the respective matrix. Using the two dimensional definition of the inverse Fourier, we can write the expectation as:

\begin{equation}
	\mathrm{E} \left[ a_{mn} \right]  = \frac{1}{MN} \sum_{m=0}^{M-1} \sum_{n=0}^{N-1} \mathrm{E} \left[ A_{kl} \right]  \mathrm{E} \left[ e^{2\pi i \left[\frac{mk}{M} + \frac{nl}{N} \right] } \right]
\end{equation}

The expectation of the exponentiation expression and $A_{kl}$ can be written as two different expectations due to there independence. Given the above expression, if the goal is to force the mean of CAI to be 0, we simply have to force the post-decomposition matrix $\mathbf{Q}$ to have an expectation of zero. Therefore as long as the $\mathbf{Q}$ matrix has a mean near zero, CAI initialization will build filters with a mean of zero.

We can also say that approximately $\mathbf{Q}$ has a expected value of $0$.

Because our matrix is symmetric, we can rewrite the decomposition as $\mathbf{Q}\mathbf{\Lambda}\mathbf{Q}^{T}$, therefore $\mathrm{E} \left[ \mathbf{Q} \right]=\mathrm{E} \left[ \mathbf{Q}^T \right]$. Because we defined as $\mathbf{A}$ as being a real positive definite matrix, all eigenvalues must be greater than 0,
therefore $0 \leq \mathrm{E} \left[ \mathbf{\Lambda} \right]$. If we naively assume covariate independence, we can rewrite our expectation as
$\mathrm{E} \left[ \mathbf{A} \right] = \mathrm{E} \left[ \mathbf{Q} \right]\mathrm{E} \left[ \mathbf{\Lambda} \right]\mathrm{E} \left[ \mathbf{Q}^T \right] = 2 \mathrm{E} \left[ \mathbf{Q} \right]\mathrm{E} \left[ \mathbf{\Lambda} \right] = 0$.
Therefore we can approximately say that $\mathrm{E} \left[ \mathbf{Q} \right] = 0$.

\subsection{Algorithm Description}

\begin{algorithm}
	\caption{2 Dimensional CAI}
	\label{CAI}
	\begin{algorithmic}[1]
		\Procedure{CAI}{$f,s,r,c, fan_{in}$}

		\State $f_{r},f_{c} \gets \mathcal{F}_{rc}$
		\State $W^{\sim} \in R^{f\times s\times f_r \times f_c}$
		\State $W \in R^{f\times s\times r \times c}$

		\For{\texttt{i from 0 to f}}
		\State $W^{\sim}_i \gets orthobasis(R^{s \times (f_r*f_c)})$
		\State $W^{\sim}_i \gets W_f \ \texttt{reshape into} \ R^{s\times f_r\times f_c}$
		\EndFor
		\For{\texttt{i from 0 to f, j from 0 to s}}
		\State $W_{i,j} \gets \mathcal{F}^{-1}\left[W^{\sim}_{i,j}\right] + \epsilon$
		\EndFor
		\State $W \gets scale(W)$
		\State \textbf{return} $W$
		\EndProcedure
	\end{algorithmic}
\end{algorithm}

The very last step of the CAI algorithm is to scale the filters variance to match the variance scheme defined in He-normal initialization. This can be done by scaling the filters by $\sqrt{\frac{2.0}{fan_{in}} / Var\left[f\right]}$  \cite{DBLP:journals/corr/HeZR015}. $\epsilon$ is random noise to break symmetry created by the inverse Fourier transform. The full description of the algorithm can be found in Algorithm~\ref{CAI}.

\section{Empirical Evaluation on Images}
\subsection{Experimental Set Up}
Convolution Aware Initialization was implemented using Theano, Tensorflow and integrated using Keras \cite{2016arXiv160502688short, tensorflow2015-whitepaper, chollet2015keras}. The algorithms were GPU accelerated on a Nvidia TitanX using Cuda 8.0 and CuDNN 5.1.
We utilized numpys implementation of the real forward and backward FFT's \cite{numpy}.

\subsection{CIFAR-10}
The CIFAR-10 dataset consists of $32 \times 32$ color images, each belonging to one of 10 classes, \cite{cifar}. The standard data-split has 50,000 training images and 10,000 test images. For data augmentation we did random horizontal flips, and cropping as described in \cite{Zagoruyko2016WRN} without $4 \times 4$ padding. We also had better results without applying any type of whitening.

The architecture chosen was wide residual network with a depth of 28, and a widening factor of $k=10$, the complete architecture description is defined at Table \ref{tab:widerecarch} \cite{Zagoruyko2016WRN}. The block type used for all residual blocks was a basic residual block without any bottleneck \cite{DBLP:journals/corr/HeZRS15}. A L2 weight decay of $0.0005$ was utilized as well.

We regularized our network with dropout as well as label-smoothing using the SoftTarget regularization scheme \cite{JMLR:v15:srivastava14a, DBLP:journals/corr/Aghajanyan16}.

We performed grid hyper-parameter optimization over dropout rate, learning rate scheduler, number of epochs trained freely as $n_b$ for SoftTarget, as well as the $\beta, \gamma$ in the SoftTarget schema \cite{DBLP:journals/corr/Aghajanyan16, Bergstra:2012:RSH:2503308.2188395}. These parameters control how much of the past soft-labels generated from the model are merged with the hard labels. Each experiment was ran 200 epochs completely over the training set. The optimization algorithm was Nesterov accelerated SGD with a learning rate of 0.01 and a momentum of 0.9. The learning rate was decreased on the schedule described by Zagoruyko et al. for their CIFAR10 experiments.

For CAI, non-convolution layers were initialized with henormal initialization \cite{DBLP:journals/corr/HeZR015}.

\begin{table}
	\centering
	\captionof{table}{Wide Residual Network Architecture}
	\begin{tabular}{lcr}
		\toprule
		group & output shape & block\\
		\midrule
		$conv_1$ & $32 \times 32$& $\begin{bmatrix}3 \times 3 & 16\end{bmatrix}$ \\
		\midrule
		$conv_2$ & $32 \times 32$ & $ \begin{bmatrix}3\times3 & 16\times k \\ 3\times3 & 16\times k\end{bmatrix} \times n$ \\
		\midrule
		$conv_3$ & $16 \times 16$ & $ \begin{bmatrix}3\times3 & 32\times k \\ 3\times3 & 32\times k\end{bmatrix} \times n$ \\
		\midrule
		$conv_4$ & $8 \times 8$   & $ \begin{bmatrix}3\times3 & 64\times k \\ 3\times3 & 64\times k\end{bmatrix} \times n$ \\
		\midrule
		$avg pool$ & $1 \times 1$ & $\begin{bmatrix}8 \times 8\end{bmatrix}$ \\
		\bottomrule
	\end{tabular}
	\label{tab:widerecarch}
	\par \bigskip
	$n$ is the amount of repetitions done per group, while $k$ represents the widening factor as described in the original paper for wide residual networks. The final layer was a densely connected layer with a softmax activation.
\end{table}

\begin{figure*}
	\centering
	\subfigure{\includegraphics[scale=0.38]{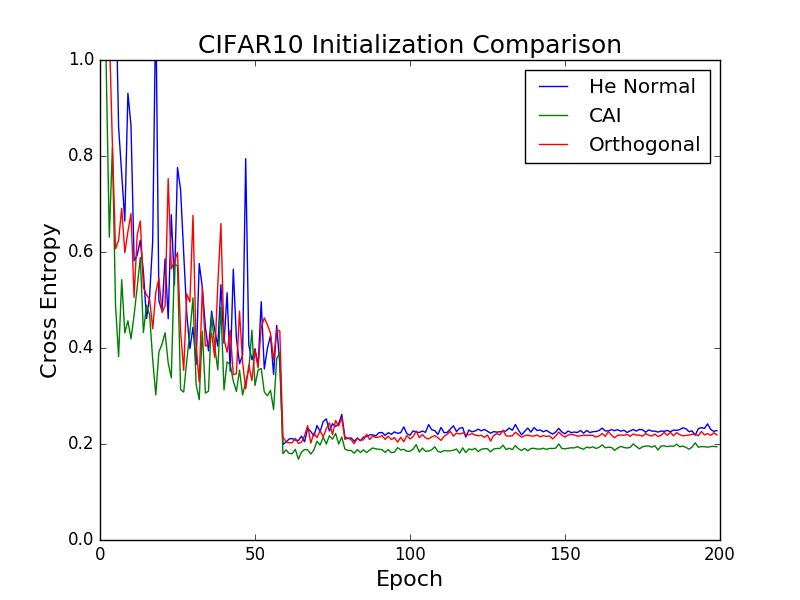}}\quad
	\subfigure{\includegraphics[scale=0.38]{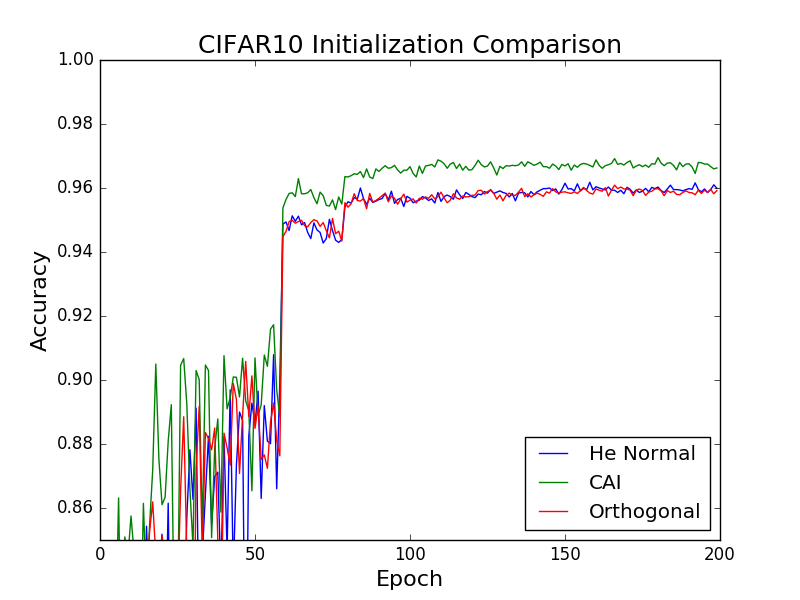}}
	\caption{CIFAR10 Validation Results}
	\label{fig:cifar10_graph}
\end{figure*}

We compare our results with the optimal results reported by Zagoruyko et al, as well as other types of initialization. To show that the increase in performance is not solely due to SoftTarget regularization we train the model without using SoftTarget as well. We report the best performing initialization using the median of 5 runs, following the experimental set-up performed by Zagoruyko et al. The reported results can be found in Table~\ref{tab:cifar_10_results} and visualization can be found in Figure~\ref{fig:cifar10_graph}.

\begin{table*}
	\centering
	\captionof{table}{CIFAR Accuracy with Data Augmentation}
	\small
	\begin{tabular}{lrr}
		\toprule
		Network                             & CIFAR10 Accuracy & CIFAR100 Accuracy \\
		\midrule
		Wide ResNet (CAI, SoftTarget)       & $\mathbf{96.31}$ & $\mathbf{79.25}$  \\
		Wide ResNet (He Normal, SoftTarget) & $\mathbf{96.18}$ & $78.31$           \\
		\midrule
		Wide ResNet (original paper)        & $96.11$          & $\mathbf{81.15}$  \\
		Wide ResNet (our tests)             & $96.12$          & $78.22$           \\
		\midrule
		Fitnet4-LSUV                        & $93.94$          & $70.04$           \\
		Fitnet4-OrthoInit                   & $93.78$          & $70.44$           \\
		Fitnet4-Highway                     & $92.46$          & $68.09$           \\
		\midrule
		ALL-CNN                             & $92.75$          & $66.29$           \\
		DSN                                 & $92.03$          & $65.43$           \\
		NiN                                 & $91.19$          & $64.32$           \\
		MIN                                 & $93.25$          & $71.14$           \\
		\midrule
		\multicolumn{2}{c}{Extreme Data Augmentation} \\
		\midrule
		Large ALL-CNN                       & $95.59$          & $68.55$           \\
		Fractional MP                       & $\mathbf{96.53}$ & $73.61$           \\
		\bottomrule
	\end{tabular}
	\label{tab:cifar_10_sota}
\end{table*}

Using Convolution Aware Initialization on residual networks reports an accuracy of $96.31$. This, as far as we know, sets a state of the art on CIFAR-10 using only basic data-augmentation such as horizontal mirroring and random shifts.
The comparison of different methods shown used data reported in
\cite{Zagoruyko2016WRN, DBLP:journals/corr/SpringenbergDBR14, DBLP:journals/corr/MishkinM15, DBLP:journals/corr/Graham14a}.

\begin{table*}

	\centering
	\captionof{table}{CIFAR-10 Results}
	\small
	\begin{tabular}{lrrrrrr}
		\toprule
		Initialization          & Validation Loss   & Accuracy         & Dropout & $n_b$ & $\beta$ & $\gamma$ \\
		CAI (SoftTarget)        & $\mathbf{0.1911}$ & $\mathbf{96.31}$ & $0.3$   & $5$   & $0.05$  & $0.5$    \\
		He Normal (SoftTarget)  & $0.1930$          & $96.18$          & $0.3$   & $10$  & $0.05$  & $0.5$    \\
		Orthogonal (SoftTarget) & $0.2008$          & $96.11$          & $0.3$   & $5$   & $0.05$  & $0.5$    \\
		\midrule
		\multicolumn{7}{c}{} \\
		\midrule
		$CAI$                   & $\mathbf{0.1920}$ & $\mathbf{96.24}$ & $0.3$   & $NA$  & $NA$    & $NA$     \\
		He Normal               & $0.1938$          & $96.10$          & $0.3$   & $NA$  & $NA$    & $NA$     \\
		Orthogonal              & $0.2028$          & $95.98$          & $0.3$   & $NA$  & $NA$    & $NA$     \\
		\bottomrule
	\end{tabular}
	\label{tab:cifar_10_results}
\end{table*}

\subsubsection{Notes of CIFAR100}
We decided to explicitly not go in depth into the CIFAR100 dataset results because the results achieved, while empirically show the benefit of CAI, did not set state of the art like CIFAR10. Therefore we opted to show CAI performance on other datasets instead. Table~\ref{tab:cifar_10_sota} shows results from CIFAR100 tests.

\subsection{SVHN}
The SVHN is a dataset of house numbers in the wild aggregated by Google \cite{Netzer2011}. The only data augmentation done was scaling the dataset by $\frac{1}{255}$. We train a wide residual network with a depth of 18 and a widening factor of 8. Refer to Table \ref{tab:widerecarch} for architecture. The architecture chosen was the one best reported by \citep{Zagoruyko2016WRN}. We also reduced learning rate automatically on plateau given a patience of 5 epochs while monitoring validation accuracy \cite{bottou-tricks-2012}, with a decay factor of $0.1$ and a minimum learning rate of $0.0005$.

\begin{figure*}
	\centering
	\subfigure{\includegraphics[scale=0.38]{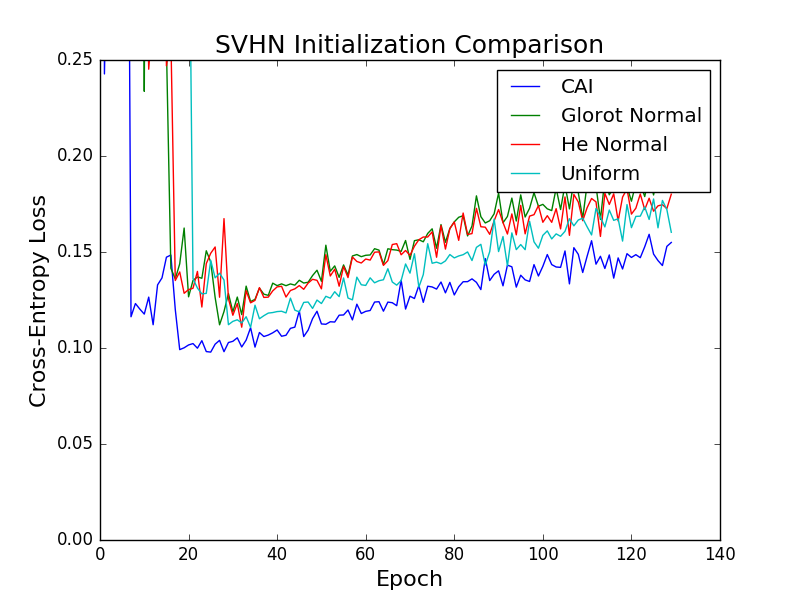}}\quad
	\subfigure{\includegraphics[scale=0.38]{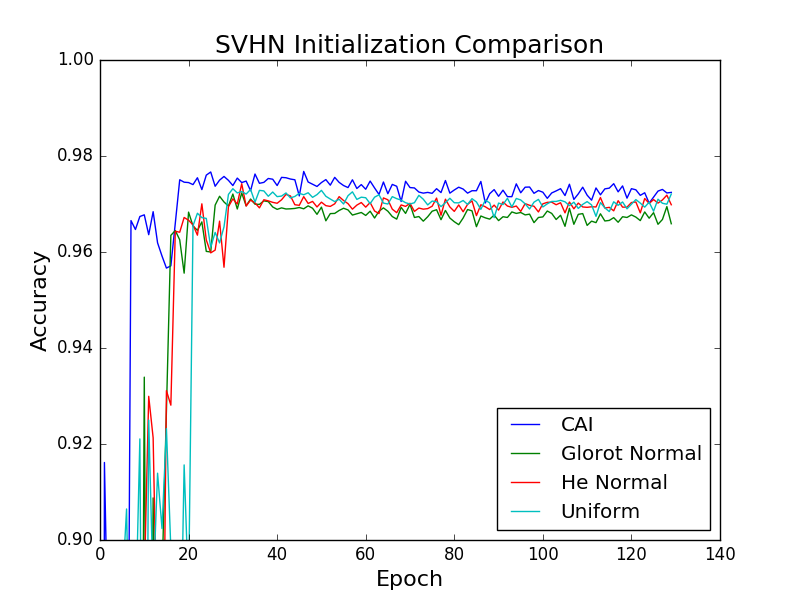}}
	\caption{SVHN Validation Results}
	\label{fig:svhngraph}
\end{figure*}

After doing hyper-parameter optimization the best dropout rate was noted to be $0.4$, and all initializations performed marginally better without the use of SoftTarget with a $L2$ weight decay of $0.0005$. We ran each experiment 5 times for 130 epochs and reported the median performing run against a set of popular initialization techniques. We report the results in Table~\ref{tab:svhn_res} and visualize the results in Figure~\ref{fig:svhngraph}.

\begin{table}
	\small
	\captionof{table}{SVHN Initialization Comparison}
	\centering
	\begin{tabular}{lrr}
		\toprule
		Initialization & Validation Loss   & Accuracy         \\
		\midrule
		CAI            & $\mathbf{0.1102}$ & $\mathbf{97.61}$ \\
		He Normal      & $0.1108$          & $97.31$          \\
		He Uniform     & $0.1223$          & $97.09$          \\
		Glorot Normal  & $0.1120$          & $97.20$          \\
		\bottomrule
	\end{tabular}
	\label{tab:svhn_res}
\end{table}

CAI peaks in performance roughly 15 epochs before all other initializations. With CAI we noticed not only higher accuracy and lower loss, but faster convergence in general.

We could not replicate the results reported by the Wide Residual Network paper in the Keras framework, even using the learning rate schedule provided by the paper \cite{Zagoruyko2016WRN}.

\section{Empirical Evaluation with 1 Dimensional CAI}
\subsection{Background}

The derivation for Convolution Aware Initialization given above was in the case of 2-dimensional convolution operator. It is also possible to derive CAI for 1 dimensional convolution. We simply have to remove the reshaping done and utilize the one dimensional forward and inverse Fourier transform. A one-dimensional implementation of CAI can be used to initialize one-dimensional convolutional layers which appear frequently in audio and NLP tasks. The algorithm description can be found at Algorithm~\ref{CAI_1d}. Our next set of experiments will empirically validate CAI for networks containing one-dimensional convolutions.

\begin{algorithm}[H]
	\caption{1 Dimensional CAI}
	\label{CAI_1d}
	\begin{algorithmic}[1]
		\Procedure{CAI}{$f,s,r, fan_{in}$}

		\State $f_{r} \gets \mathcal{F}_{r}$
		\State $W^{\sim} \in R^{f\times s\times f_r}$
		\State $W \in R^{f\times s\times r}$

		\For{\texttt{i from 0 to f}}
		\State $W^{\sim}_i \gets orthobasis(R^{s \times (f_r*f_c)})$
		\State $W^{\sim}_i \gets W_f \ \texttt{reshape into} \ R^{s\times f_r}$
		\EndFor
		\For{\texttt{i from 0 to f}}
		\State $W_{i} \gets \mathcal{F}^{-1}\left[W^{\sim}_{i}\right] + \epsilon$
		\EndFor
		\State $W \gets scale(W)$
		\State \textbf{return} $W$
		\EndProcedure
	\end{algorithmic}
\end{algorithm}

\subsection{IMDB Movie Review}
The IMDB Movie Review dataset, is a sentiment analysis dataset containing 25,000 movie reviews tagged by sentiment; positive or negative. We focus testing on a standard architectures that utilize some mixture of embedding, one-dimensional convolution, and recurrent neural layers \cite{Gal2015Theoretically,Hochreiter:1997:LSM:1246443.1246450}.
For our recurrent network we chose to use a LSTM network. LSTM networks have been used extensively successfully in various NLP tasks due to there ability to learn patterns between long time periods \cite{DBLP:journals/corr/abs-1211-5063, hong2015sentiment}.

For preprocessing we filter out a subset of words for the sentences and store each sentence as a matrix where individual rows represent a word via one hot encoding. We pad each sentence to insure that each matrix of a sentence was the same size as every other sentence. We used a maximum of 20,000 unique words and limited sentences to a maximum length of 80. We used the standard binary cross-entropy loss. Three architectures were tested:

\begin{itemize}
	\item Embedding $\rightarrow$ LSTM $\rightarrow$ Dense
	\item Embedding $\rightarrow$ Convolution1D $\rightarrow$ GlobalPoooling1D $\rightarrow$ Dense
	\item Embedding $\rightarrow$ Convolution1D $\rightarrow$ Poooling1D $\rightarrow$ LSTM $\rightarrow$ Dense
\end{itemize}

The hyper-parameters in all three models as well as the hyper-parameters of the Adam optimization method were chosen using random hyper-parameter search \cite{Bergstra:2012:RSH:2503308.2188395, DBLP:journals/corr/KingmaB14}. Every configuration was run 5 times, and the median run was reported. For CAI, all non-convolutional layers were initialized with orthogonal matrices. The results can be found at Table~\ref{tab:imdb}.

Once again CAI outperformed all other forms of initialization.

\begin{table}
	\centering
	\captionof{table}{IMDB Movie Review Architecture+Init Results}
	\small
	\begin{tabular}{lr}
		\toprule
		Initialization                   & Accuracy         \\
		\midrule
		\multicolumn{2}{c}{Embed-LSTM} \\
		\midrule
		Orthogonal($scale=0.3$)          & $\mathbf{90.02}$ \\
		Uniform($low=-0.05, high= 0.05$) & $89.78$          \\
		Normal($\mu=0, \sigma=0.3$)      & $89.00$          \\
		CAI                              & $NA$             \\
		\midrule
		\multicolumn{2}{c}{Embed-Conv} \\
		\midrule
		Orthogonal($scale=0.3$)          & $89.63$          \\
		Uniform($low=-0.05, high= 0.05$) & $89.20$          \\
		Normal($\mu=0, \sigma=0.3$)      & $89.18$          \\
		CAI                              & $\mathbf{90.88}$ \\
		\midrule
		\multicolumn{2}{c}{Embed-Conv-LSTM} \\
		\midrule
		Orthogonal($scale=0.3$)          & $90.31$          \\
		Uniform($low=-0.05, high= 0.05$) & $89.78$          \\
		Normal($\mu=0, \sigma=0.3$)      & $90.16$          \\
		CAI                              & $\mathbf{91.40}$ \\
		\bottomrule
	\end{tabular}
	\label{tab:imdb}
\end{table}

\subsection{Speech Synthesis via WaveNet}
The next experiment we ran was using the wave net architecture to perform speech synthesis trained on the VNTK dataset \cite{DBLP:journals/corr/OordDZSVGKSK16, veaux2016cstr}. The reason we decided to run this specific experiment was to test CAI with stacked one dimensional convolutional layers, as well as to see how CAI performs with atrous convolutions \cite{chen2016deeplab}.

\begin{figure}
	\centering
	\includegraphics[scale=0.38]{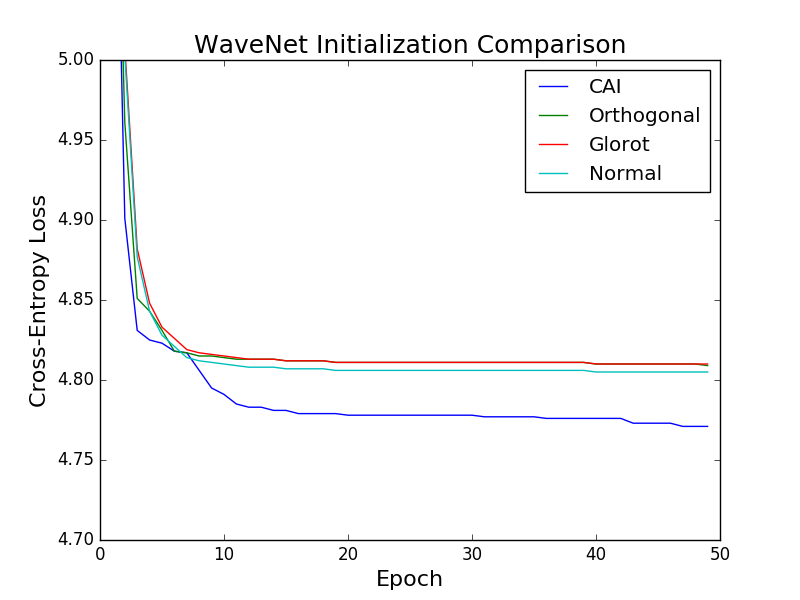}
	\caption{WaveNet Validation Results}
	\label{fig:wavegraph}
\end{figure}

We trained a small version of the wavenet architecture, which contains a sample rate of 4000, with 256 output bins. We utilized skip connections as originally proposed in the paper, with 256 filters for every convolutional layer, with a dilation depth of 9 for the dilated or atrous convolution layers. We also did not use any bias additions in the networks. We used Nesterov accelerated stochastic gradient descent with a learning rate of $0.1$ and a momentum rate of $0.9$. We trained the network for 50 epochs \cite{DBLP:journals/corr/OordDZSVGKSK16}. We ran this set of experiment 5 times and reported the median run. Results are shown in Table~\ref{tab:wavenet} and Figure~\ref{fig:wavegraph}.

CAI outperformed other standard schemas of initializations by a wide margin.

\begin{table}[H]
	\centering
	\captionof{table}{WaveNet Initialization Results}
	\small
	\begin{tabular}{lr}
		\toprule
		Initialization              & Categorical Cross Entropy \\
		\midrule
		Orthogonal($scale=0.2$)     & $4.809$                   \\
		Glorot                      & $4.810$                   \\
		Normal($\mu=0, \sigma=0.2$) & $4.811$                   \\
		CAI                         & $\mathbf{4.771}$          \\
		\bottomrule
	\end{tabular}
	\label{tab:wavenet}
\end{table}

\section{Discussion}
In this paper we introduced a new type of initialization which takes into account the properties of convolution. We showed reasoning behind building orthogonal basis in the Fourier space rather in the standard space, showing that convolution across a stack is similar to a linear combination of filters in the Fourier space. The paper also proved the bounds of eigen-decomposition on a random symmetric matrix as well as the bounds of CAI prior to scaling. We also proved the preconditions necessary to force the mean of CAI to zero.

From an empirical testing perspective, CAI outperformed other standard types of initialization across the board, setting a new state of the art for the CIFAR10 dataset with basic data-augmentation. On other tasks CAI networks converged significantly faster than other standard forms of initialization.

Further work can explore variance scaling schemes other than He Normal, and extending the idea of CAI to a mix of recurrent and image convolutional networks.

\bibliography{example_paper}
\bibliographystyle{icml2017}

\end{document}